%% file: main.tex
\documentclass[10pt,conference]{IEEEtran}
\IEEEoverridecommandlockouts 
\usepackage{times}  
\usepackage{helvet}  
\usepackage{courier}  
\usepackage{url}  
\usepackage{graphicx}  
\usepackage{booktabs}  
\usepackage{mathtools} 
\usepackage[linesnumbered, boxed]{algorithm2e}

\usepackage{array}
\usepackage{caption}
\usepackage{amsthm}
\usepackage{authblk}
\newcommand{\hd}{\rotatebox{60}}
\graphicspath{ {./figures/} }
\usepackage{color, colortbl}

\title{\LARGE \bf Automatically detecting data drift in machine learning classifiers}

\author[1]{Samuel Ackerman}
\author[1]{Orna Raz}
\author[1]{Marcel Zalmanovici}
\author[1]{Aviad Zlotnick}
\affil[1]{IBM Research, Israel}

\begin{document}
\maketitle
\input{abstract.tex}
\input{intro.tex}

\input{methodology.tex}

\input{evaluation.tex}
\input{related.tex}
\input{discussion.tex}
\bibliographystyle{abbrv}
\bibliography{main} 
\end{document}

%% file: abstract.tex
\begin{abstract}
Classifiers and other statistics-based machine learning (ML) techniques generalize, or learn, based on various statistical properties of the training data. The assumption underlying statistical ML resulting in theoretical or empirical performance guarantees is that the distribution of the training data is representative of the production data distribution.

This assumption often breaks. Statistical characteristics of the data may change between training time and production time, as well as throughout production. For example, sensors used to acquire data may lose sensitivity with time, populations may age, and fashion trends may change.

These changes might affect the ML performance, such as lower a classifier accuracy below an anticipated threshold. In some cases, such as medical applications, lower accuracy may be crucial. We term changes that affect ML performance `data drift' or `drift'. 

Many classification techniques compute a measure of confidence in their results. This measure might not reflect the actual ML performance \cite{guo2017calibration}. A famous example is the Panda picture that is correctly classified as such with a confidence of about 60\%, but when noise is added it is incorrectly classified as a Gibbon with a confidence of above 99\% \cite{pandagibbon}. However, the work we report on here suggests that a classifier's measure of confidence can be used for the purpose of detecting data drift.

We propose an approach based solely on classifier suggested labels and its confidence in them, for alerting on data distribution or feature space changes that are likely to cause data drift. Our approach identities degradation in model performance and does not require labeling of data in production which is often lacking or delayed.  Our experiments with three different data sets and classifiers demonstrate the effectiveness of this approach in detecting data drift. This is especially encouraging as the classification itself may or may not be correct and no model input data is required. We further explore the statistical approach of sequential change-point tests to automatically determine the amount of data needed in order to identify drift while controlling the false
positive rate (a.k.a Type-1 error).
\end{abstract}

%% file: intro.tex
\section{Introduction}\label{sec:intro}
This paper addresses the issue of detecting when the performance, such as accuracy, of an operational ML-based system degrades and becomes unacceptable.

How can this happen? "Panta Rhei!" (Ever newer waters flow on those who step into the same river, Heraclitus). Consider a model trained to detect images of a cell phone prior to the introduction of smart phones. Even if the model had 100\% accuracy, it would perform miserably on modern day images where virtually all cellphones are smart phones with a huge screen and no keyboard. Obviously, the transition from dumb phones to smart phones was gradual, and one would like to know that the ML system is performing poorly as early as possible. 

Another case where accuracy can degrade is when sensors used for capturing data, such as, an  X-Ray lamp, age, and properties of captured data change. In this case, performance degradation might be life threatening.

More formally: there is often a gap between ML model performance in training, which is acceptable, and model performance in production, which may drift away and become unacceptable. 
ML techniques, such as regression, classification and clustering, make statistical assumptions in order to generalize or `learn' from training data examples. 
Often, the statistical characteristics of the data change between training time and production time, as well as throughout production. 
These changes might affect the ML performance, such as lower a classifier accuracy below a required threshold.


We refer to the gradual change in input data that impacts ML model performance as \emph{Data Drift}, and claim that detection of such drift is a significant aspect of system ML quality. Apparently, there is a lack of both methodology and technology supporting the assessment of this quality aspect in ML-based systems. Many common approaches, such as remeasuring ML performance in production, are not feasible when production labels are unavailable or delayed. 

It is important to note the following two challenges.
 \\(1) A change in distribution of classes (such as, previously there were more dogs than cats, but now there are more cats than dogs) is not data drift, and, normally, would not impact system accuracy. Furthermore, comparing distributions, e.g., between training or testing and production data sets, is not feasible because the data distribution may be unknown. The challenge of distribution identification is a well-acknowledged one, for example underlying much of the Bayesian networks work, and is one of the challenges that we address in our research. Density estimation is largely unfeasible because of the high dimensionality of the data, though there are attempts to make progress on that front (see Section Related).
\\(2) Frequently comparing production output to true labels may be unfeasible. Labeling may be expensive, experts able to label the input  may be unavailable, and during production labeling is frequently delayed, which may be intolerable.
    We note that while there is substantial experience in developing successful ML solutions in areas where there is a lot of labeled training data, labeling is cheap, and only light domain knowledge is required, less experience exists in developing successful ML solutions when any of these conditions break. However, these conditions often break in the domain of business applications. 


The main contributions of this work are (1) defining the problem of quality under drift, (2) defining the notion of an \emph{auditor}, a detector of drift, introducing an example auditor, the classifier confidence auditor, (3) a method to introduce drift for testing without impacting the data semantics, (4) a framework to simulate drift and assess the effectiveness of auditors in detecting it, and (5) demonstrating how the statistical approach of sequential changepoint analysis may be applied to the challenge of drift detection such that the false positive error rate can be controlled in a statistical sound manner.


The Methodology Section
introduces the classifier confidence auditor, drift types, and drift detection framework. These are used in our experiments to demonstrate the effectiveness of the classifier confidence auditor in detecting drift over three different data sets and various classifier models, as reported in the Experimental Results Section.
Next, Related work is discussed. 
The Discussion Section
concludes the paper.

%% file: methodology.tex
\section{Methodology}\label{sec:methodology}
We provide information on the classifier confidence drift detection algorithm, the data sets and classifier models that we use in our experiments, the experimental settings, and our drift detection framework, concentrating on drift simulation and drift types. 

\subsection{The Classifier confidence drift detection algorithm} \label{subsec:algorithm}

We show that if production class confidence values do not exhibit statistical similarity to baseline  class confidence values (defined below) there is data drift, meaning that the production data has deviated significantly from training data such that classification results may be incorrect.

To take advantage of the strength of classical statistical techniques, e.g., for comparing the identity of two empirical distributions, we define a uni-variate statistic over the given high dimensional non-parametric empirical data distribution. In the case of the classifier confidence auditor this statistic is the distribution of the classifier's own labeling confidence. 
The classifier assigns a label for every input data. This is the 'winning' label. This assignment is based on a statistical measure of confidence. 
The auditor works over two distributional sources of variability. One is the distribution of the classifier confidence in its label assignment for a data record $d$, $P_{\text{conf}}(d)$. We term this the `winning label confidence distribution' or `label-agnostic confidence distribution' since it depends on the data record and not the label. Another is the distribution of the classifier confidence in its per-label $c$ assignment when a data record $d$ is assigned label $c$, $P_{\text{conf}}(d|c)$. We term this a `per-label confidence distribution'. Notice that there are multiple per-label confidence distributions, one per label/class.

The drift detection algorithm is summarized in Algorithm 1.
One data set is indicated as a baseline, for example, the test data set. This means that the classifier operation was determined to be satisfactory on it and that the data set was determined to be representative of the confidence distribution expected in production. The drift detection algorithm performs a non-parametric statistical test for the identity of two uni-variate distributions over the classifier confidence distributions from the two data sets. Examples of such statistical tests are the two-sample two-tailed T-test and the two-sample Kolmogorov-Smirnov test.
A drift alert is triggered if the distributions are significantly different.

\begin{figure}[ht]
\centering
\includegraphics[width=0.46\textwidth]{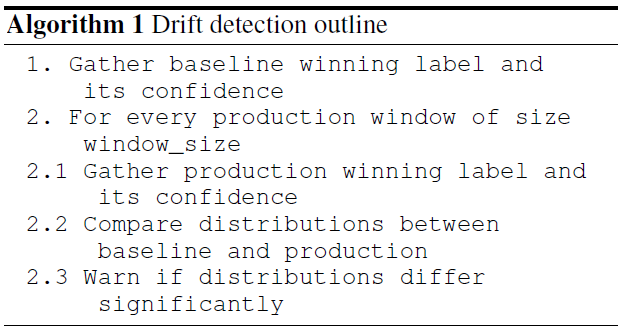}
\label{fig:pseudocode}
\end{figure}

\begin{figure}[ht]
\caption{Small drift, undetected by distribution test}
\label{fig:undetectedDrift}
\centering
\includegraphics[width=0.45\textwidth]{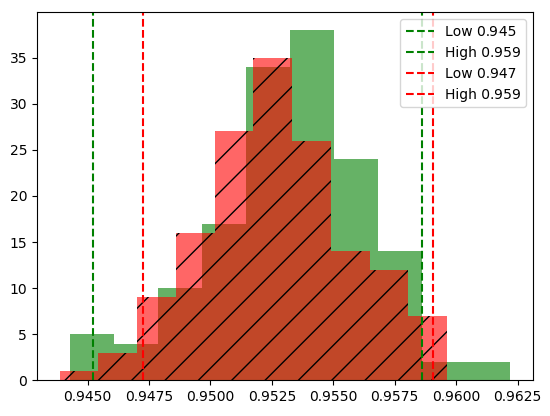}
\end{figure}

\begin{figure}[ht]
\caption{Larger drift, detected by distribution test}
\label{fig:detectedDrift}
\centering
\includegraphics[width=0.45\textwidth]{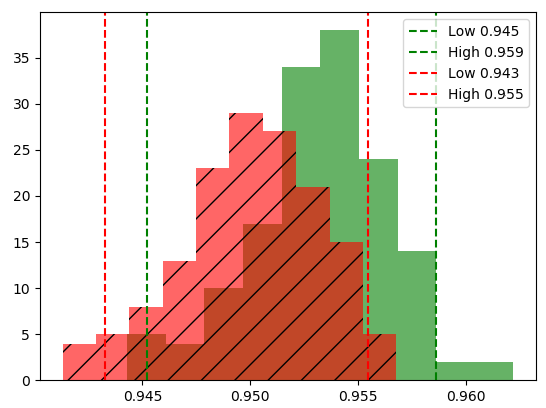}
\end{figure}

Figures \ref{fig:undetectedDrift} and \ref{fig:detectedDrift} show an example of drift introduction over MNIST data.
We train a classifier without one of the digits (digit '3'). Then, simulating production drift, we inject a slowly increasing percentage of the held-out digit into the data to be classified (adding images of digit '3'). The figures show the confidence intervals over the classifier confidence for the winning label. Green (untextured) is the original data (without '3'), red (textured) is the 'production' data (which includes some  digit '3' images). The statistical test (two-sample two-tailed T-test in this case) is not able to identify the low percentage drift (Figure \ref{fig:undetectedDrift}) but is able to clearly identify the larger percentage drift (Figure  \ref{fig:detectedDrift}).

\subsection{Data and classifiers}\label{subsec:data}
We validate our classifier confidence drift detector on three different data sets and multiple classifiers. Following is a description of the three data sets being used in our experiments and the classifiers that were used for each one. Notice that the approach is agnostic to the type of classifier trained. All it requires is the classifier's confidence in the winning label. The approach is also agnostic to the type of data the classifier works on. This is because it does not require any training or production input data. It requires only classifier output information as stated above. 

\begin{itemize}
    \item MNIST \cite{MNIST_data} ("Modified National Institute of Standards and Technology") is the de facto “hello world” dataset of computer vision. MNIST is a data set for handwritten images (digits) which has served as the basis for bench-marking classification algorithms since its release in 1999.
    MNIST contains 70,000 gray-scale 28x28 images (out of which 10,000 are used for testing) of digits divided into 10 classes (one per digit). The classifier task is to classify an incoming image into one of the 10 digit classes.
    We trained Logistic Regression and CNN models on this data.
    \item CIFAR10 \cite{CIFAR10_data} is an established computer-vision data set used mainly for object recognition. It is a subset of the "tiny images data-set" and consists of 60,000 32x32 color images (out of which 10,000 are used for testing) containing one of 10 object classes (e.g. dog, car), with 6,000 images per class.
    We trained Logistic Regression (not reported on due to low accuracy) and CNN models on this data.
    \item Kaggle lending club loan data \cite{Kaggle_loan_club_data} contains the complete loan data for loans issued 2007-2015, including the loan status and latest payment information. It consists of 887k records with 75 features (30 of them numeric).
    We trained Random Forest classifiers on this data. 
\end{itemize}

\subsection{Experiment design} \label{subsec:design}
When applying the classifier confidence auditor, one only needs to provide the classifier's confidence per label, or even just per the winning label, for each data record in the data set under investigation. 
In production, the classifier model is fixed and provided by the user. However, for assessing auditors effectiveness, our framework repeatedly \textit{simulates} different types and amounts of drift.
In some of our experiments we train simulation-specific classifier models such that we can control the drift type and amount. See Section Drift Simulation
for details. 

For each data set and model, our auditors framework first creates a baseline for the statistic under investigation over the baseline data set. In the case of classifier confidence auditor, the statistic is the winning-label confidence. In our experiments we also compare against per-label confidence.

The framework then runs simulations in which it compares the baseline statistic distribution with that of the same statistic over the data set simulating the production drift.
The framework introduces increasing amount of drift into the simulated production data set. While doing so, the framework retains the original ratio of data records per label for the no-drift part of the data set. This is done in order to prevent unintended change in the original data distribution. 

The goal is to find the minimal level of drift in the data that the auditor consistently identifies. Consistency is determined by a series of at least 50 simulation iterations per the the same drift level, where at least half of the iterations detect drift.  
Section Experimental Results 
provides the experimental results and their evaluation.

\subsection{Drift simulation} \label{subsec:drift}
We define drift as changes in the data that affect the model ML performance. We are particularly interested in changes that cause a classifier accuracy to drop below the expected interval lower threshold. Even getting above the top of the expected interval (doing too good) is an interesting observation.

Introducing drift is a non trivial task. For example, it is challenging to know what the correct label is, in case of changing the data. In addition, we need to be careful not to change the semantics of the underlying data. Changing data might even result in data records that are illegal or impossible. 
On the other hand, real drift is rarely available for experimentation. Usually, if such data is available it is used for training and thus its information is already included in the model. Therefore, we found it necessary to incorporate strong drift-creation heuristics into our auditors framework. 

The framework supports the definition of new types of drift per auditor. We have implemented six general types of drift, as detailed below. In our experiments we mostly use drift types 1--4, as these types ensure that all data is legal and representative of the expected distribution. By definition, types 1--4 cause the classifier to provide an incorrect classification, thus introducing measurable drift. 
\begin{enumerate}
    \item Data labels. A model is trained over a subset of data records, such that all data records for a particular label(s) are left out of the training data. The left-out data records are then randomly selected as drift. This type of drift does not make any changes to the data thus preserving semantics, which is a desirable property. This type of drift relies on having labels.  
    \item Data records per selection criteria. This is similar to type 1 drift, with the exception of providing selection criteria for the left out data records, rather than relying on labels. An example criterion is the top 10th percentile of data records as ordered by a particular feature's values. Type 2 drift does not necessarily cause data drift.
    \item Out-of-domain data. This is data that is given in a legal format yet has different semantics. For example, a gray-scale cat image given to a gray-scale digit classifier.
    \item Random data. This is data that is given in a legal format yet has random values. For example, random gray-scaled pixels in an image. There is no need to label the random data as it has no legal class. 
    \item Feature changes per given function. For example, multiply a feature values by a constant. While it is easy to control the properties of this drift type, it might become unclear what the label of a resulting data record is. 
    \item Data transformation. Symmetry learning and group representation theory, e.g., \cite{NIPS2014_5424,Kiddon2015SymmetryBasedSP}, provide us with the theoretical and often practical foundation to transform the data in certain domains such that its labels remain invariant. An example is rotating animal images. A rotated image of a dog retains the label of 'dog'. Ideally, such transformations should already be included in the training data. In practice, however, this is often not the case. 
\end{enumerate}

Notice that because drift types 1 and 2 define data records from the original data set as drift, the simulation includes training simulation-specific models. In type 1 drift the auditors framework trains a model $M(D  \setminus D_c)$ over the original data $D$ holding out data $D_c$ that is labeled as the drift class $c$. $D_c$ is then inserted into the production data in slowly increasing percentages as drift. This is done for each of the classes, training a model per held-out label. Similarly, in type 2 a model is trained without the held-out data records. 

\begin{figure}[ht]
\caption{Winning class and confidence at different drift \%}
\label{fig:confidence_drift}
\centering
\includegraphics[width=0.45\textwidth]{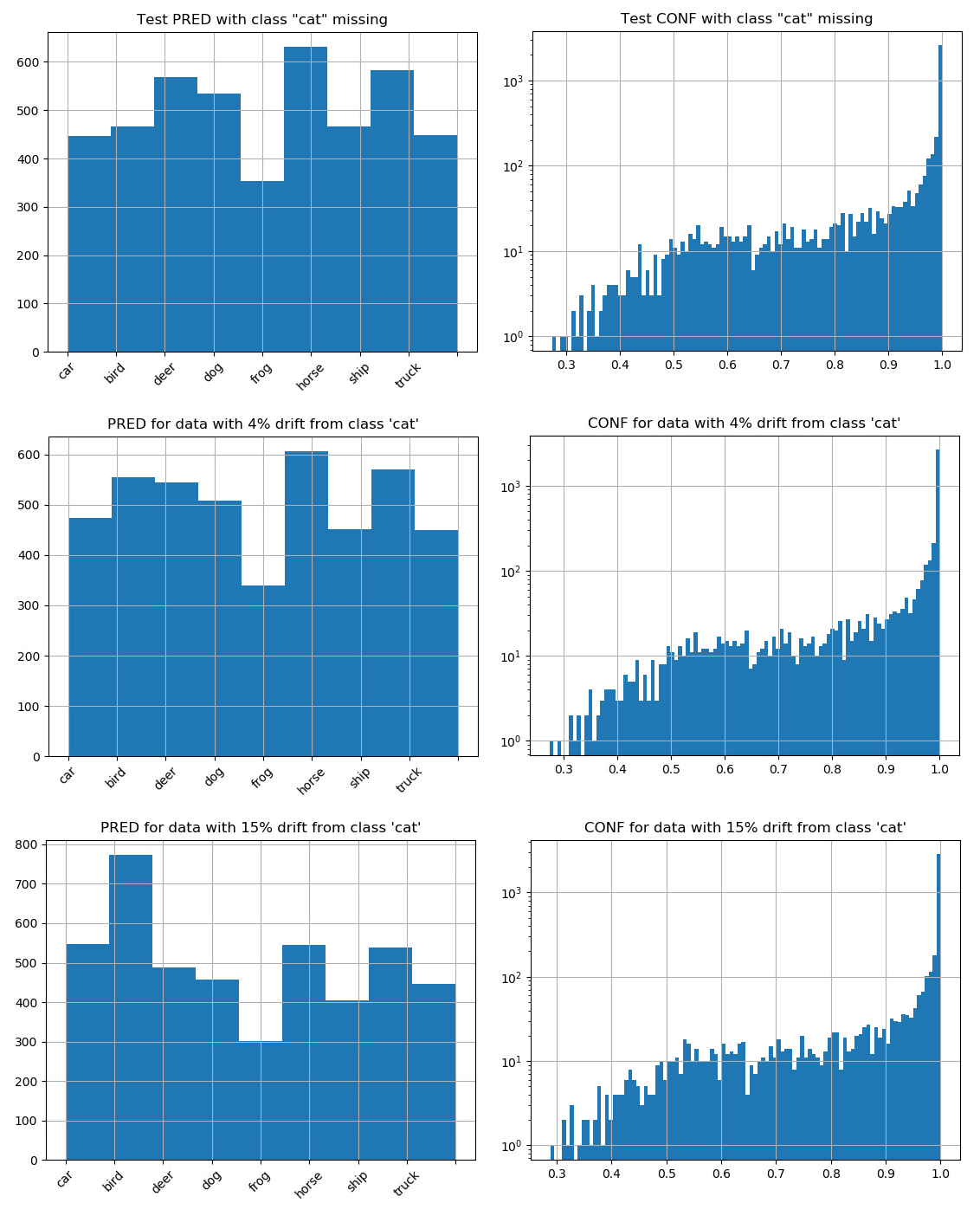}
\end{figure}

\begin{figure}[ht]
\caption{Winning class and confidence for drift (class 'cat') data}
\label{fig:confidence_only_drift}
\centering
\includegraphics[width=0.45\textwidth]{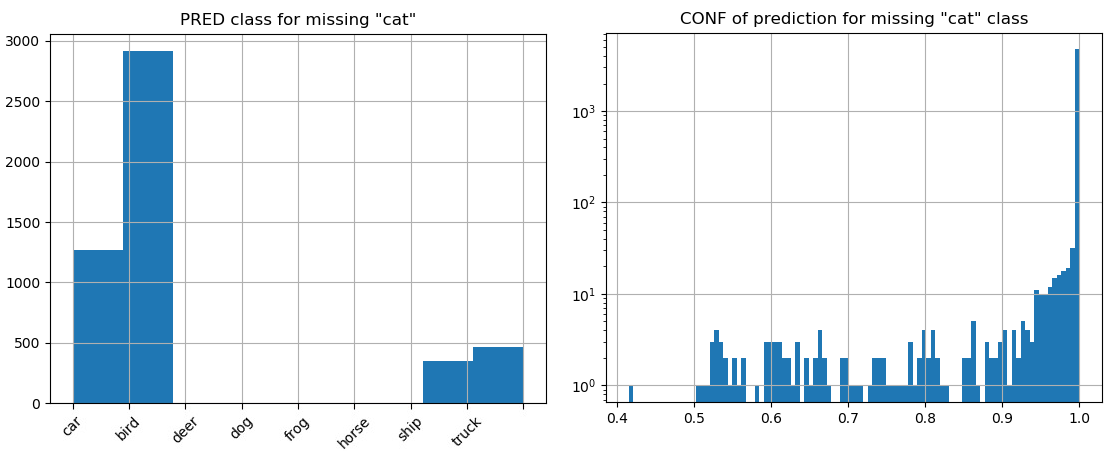}
\end{figure}

The histograms in 
Figures \ref{fig:confidence_drift} and \ref{fig:confidence_only_drift} provide examples of the distribution (Y axis is count) of the winning class (left) and of the winning class confidence (right).
The different figures (top to bottom) demonstrate the effect of inserting different amounts of drift. In Figure \ref{fig:confidence_drift} the top images depict this information over data without drift. One can see that the class distribution is not uniform. The middle images show little difference in both distributions when inserting 4\% drift. However, as the drift increases, as shown in the bottom images with 15\% drift, we can see a shift both in the class distribution (in this case more images are classified as 'car') and in the overall confidence in the winning classification. Notice that this confidence may be lower or higher than the original confidence. In this example the classifier confidence auditor identifies drift starting from 7\% type 1 drift. 
Figure \ref{fig:confidence_only_drift} shows, for reference, the class distribution and winning label confidence in the extreme case where only drift exists.

\subsection{Sequential Change-point analysis} \label{subsec:changept}

Sequential change-point methods are a very general tool to detect when a sequence of observed data $x_1,x_2,\dots$ undergo a change in distribution.  Typically, such methods can statistically determine a time of change $k$ such that samples $\{x_1,\dots,x_k\}$ and $\{x_{k+1},\dots,x_t\}$ appear to come from different distributions $F_1$ and $F_2$, respectively, where $k$ may maximize the statistical distance between the samples. This corresponds to a determination at time $t$ that drift began at time $k<t$.  The advantage of such methods is that the size of the production data used to detect drift does not need to be pre-specified, and that the drift determination can be made at a time $t$ without observing data after this. 

Importantly, such sequential methods are designed to offer a statistical guarantee on the drift detection, such as controlling the false positive rate (Type-1 error probability), that is, the probability of falsely detecting drift when it has not happened, while allowing change to be detected at any time $t$.  This is statistically difficult because with traditional hypothesis tests, like a T-test, the sample size $n$ must be pre-fixed and the data cannot be examined before $n$ for the statistical guarantees to remain valid.  Unfortunately, many such methods such as CUSUM rely on likelihood ratio tests which require the distributions $F_1$ and $F_2$ that we consider as the pre-drift and drift distributions to be known, which is a weakness.

We will use a very flexible approach to drift detection in univariate data, presented in \cite{cpm_article}, and implemented as the \texttt{R} package \texttt{cpm} (change point models, \cite{cpm_package}).  These CPM methods have the combination of advantages of (1) not specifying the distributions $F_1$ and $F_2$ that we test drift on; (2) allowing several different statistical tests to detect different types of distributional change, such as T-test, Cramer-von-Mises (CvM), and Kolmogorof-Smirnov; and (3) being able to retroactively detect drift at any time $t$ while maintaining the false positive rate at a fixed $\alpha$ at any time $t$, conditional on not having detected drift prior to $t$.  These CPM methods will be applied on sequences of univariate data of the estimated class probability of the class that is predicted.


%% file: evaluation.tex
\section{Experimental results}
\label{sec:eval}
To assess the effectiveness of the classifier confidence auditor for drift detection we experiment with the data sets described in Section Data and classifiers.
Our auditors framework runs drift simulations, testing multiple drift types, as described in Section Drift simulation.


The simulation takes a data set and defines a training data set, a baseline data set, and multiple production data sets. The simulated production data is composed of data that was used neither for training nor for creating the baseline. The simulated production data is a mixture of data coming from the same distribution as the training data with varying percentages of drift data. In the experimental results we indicate not only the drift type but also the drift percentage. 

We are interested in answering the following research questions:
\newtheorem{question}{RQ}
\begin{question}
Is there a minimal drift percentage from which the auditor consistently detects drift? In other words, what is the drift-detection ability of an auditor?
\end{question}
\begin{question}
Is the drift-detection ability affected by the drift type? 
\end{question}
\begin{question}
How does a label-agnostic auditor compare to a per-label auditor?
\end{question}
\begin{question}
Does the ML performance of a classifier affect the drift-detection ability?
\end{question}
\begin{question}
Can an auditor based solely on label distribution have an acceptable drift-detection ability?
\end{question}
\begin{question}
Can change-point analysis provide statistical guarantees to further improve drift detection?
\end{question}{}
Following are results that answer our research questions. 
\subsection{RQ 1: Drift-detection ability} \label{subsec:detectionAbility}
Table \ref{tab:experimentSummary} summarizes the experimental results in terms of drift-detection rate. Each row summarizes all the experiments on the specified data set, type of classifier model trained and type of drift inserted. Except for drift type 2 (held-out feature subset) each row summarizes simulations over multiple different models. For each class (10 in MNIST and CIFAR10, 7 in loan data) a model is trained over data without the class, while drift inserts data records from that class in increasing percentages into the simulated production data set. Tables \ref{tab:mnistlogreg} to \ref{tab:cifar10NN} summarize the results per model, indicating the left-out class inserted as drift. The tables show that for a combination of data set, classifier model and drift type there is a drift percentage for which the auditor consistently detects drift. As expected, differences in the data, model and drift type affect the drift-detection ability. 
The ability of the auditors to detect drift above a threshold is demonstrated, via simulations, not only by the difference among the drift
types but also by the variability among the models
created for the same drift type over different data subsets.
The 'Min \%' column indicates the model and drift type for which the drift-detection ability was the best. Similarly, the 'Max \%' column indicates the model and drift type for which the drift-detection ability was the worst. The 'Avg \%' column shows the average drift-detection ability over all the models.

\subsection{RQ 2: Effect of drift type on drift-detection ability}\label{subsec:affectOfType}
Drift type affects drift-detection ability.
Random input seems more difficult to detect than semantically meaningful input, even if it comes from a different domain, such as inserting an adjusted CIFAR10 image into MNIST classifier.

We observe that type 1 drift (held-out class) often causes high classifier confidence in its incorrect labeling. A possible explanation is that the model has a 'default' class i.e., one chosen when no other clear winner exist, and the drift records are mostly classified as that class. If the model confidence $p(c|d)$ of a class $c$ given a data record $d$, can be broken additively into $p(c|d) = p_{\text{learned}}(c|d) + p_{\text{prior}}(c|d)$, the learned and prior confidence respectively, then possibly under the new data record $d$, the $p_{\text{learned}}(c|d)$ are similar across all original classes. Thus, a classifier based on Bayes rule will choose according to the priors.  If the priors are biased towards the default class then the new class will be classified as default with high confidence.

Alternatively, it may be the case that each class is being generated from some distribution over $R^n$, e.g., a mixture of Gaussians where each Gaussian is a class. 
The drift class will be labeled as the default class if the average and most of the probability mass of the drift class fall inside the decision boundaries of the default class.

As observed in RQ 2, type 4 drift (random) is often more difficult to detect than a drift type that has semantics (such as type 1 and 3). 
This may be because our implementation for random drift creates perturbations that are topologically close to the training data.  Given classifiers that are continuous or partially continuous (such as neural networks that exploit step functions), such perturbations are more likely to be well within the decision boundaries than very close to their threshold. In contrast, drift 1 or 3 classes are most probably not close topologically to the original classes and will thus be easier to identify.  Naturally, other factors can intervene with this intuition such as including the random data records in the training set.      


\subsection{RQ 3: Classifier confidence auditor that is agnostic to predicted labels}
\begin{figure}[ht]
\caption{MNIST CNN per class drift-detection heatmap}
\label{fig:heatmap}
\centering
\includegraphics[width=0.45\textwidth]{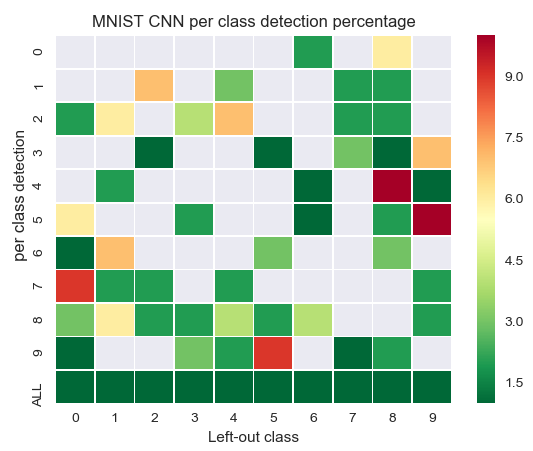}
\captionsetup{width=.95\columnwidth}  
\caption*{X axis specifies the left-out class (one of the digits 0--9). Y axis is the per-label drift-detection indication. The bottom 'All' row summarizes the winning label (label agnostic) detection, as opposed to the other rows which show per-label detection. The color of the cell indicates the minimum amount of drift required for detection, per the legend on the right. Bottom/green is small, top/red is larger (but no more than 10\%). Uncolored cells indicate a drift-detection value of above 10\%.
 }
\end{figure}

\begin{figure}[ht]
\caption{CIFAR CNN per class drift-detection}
\label{fig:classvsall}
\centering
\includegraphics[width=0.45\textwidth]{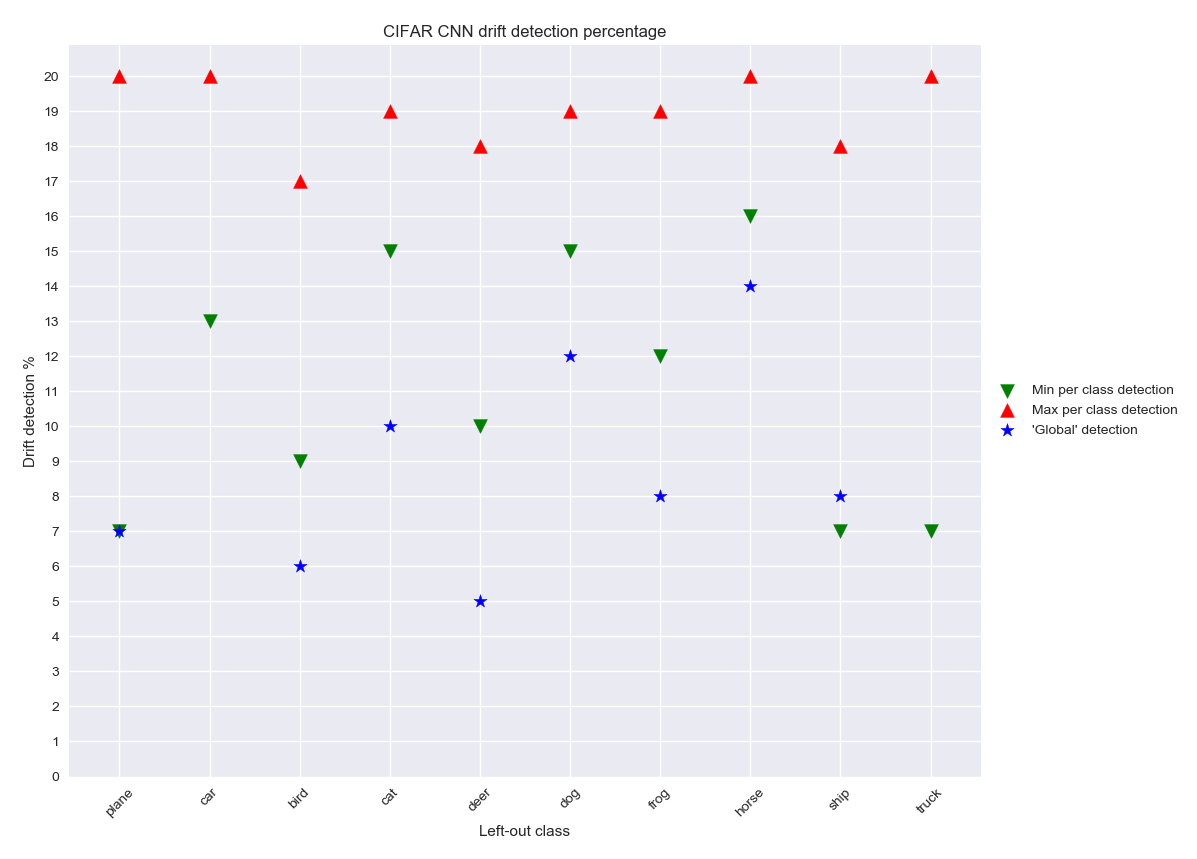}
\captionsetup{width=.95\columnwidth}  
\caption*{The Figure displays the same information as Figure \ref{fig:heatmap} for a different data set, visualized differently. X axis is the left-out class model (a model per left-out class). Y axis is the simulation drift percentage introduced. The red/upper triangle icons mark the latest drift-detection value for any of the per-label auditors. Similarly, the green/down triangle icons mark the earliest drift-detection value for any per-label auditors. The blue/star icons show the global/'All' label agnostic drift-detection value when comparing winning-label distributions. }
\end{figure}

Figures \ref{fig:heatmap} and \ref{fig:classvsall} depict the effect on the drift-detection ability of comparing the winning-label (label agnostic) confidence distribution to that of the per-label confidence distributions. The former tests for a difference in the distribution of the confidence in the winning-label regardless of its assigned class, while the latter tests for a single label that shows significant difference in the distribution of confidences. 

Both figures depict the results of the simulation on one of the data sets with one of the drift types but are representative of the results in general. The lowest row in Figure \ref{fig:heatmap} shows that the label-agnostic auditor is more reliable than the per-label auditors. Figure \ref{fig:classvsall} shows that it is usually the case that the label-agnostic auditor is comparable to or better than the best per-label auditor. Moreover, there are often cases in which none of the per-label auditors trigger, while the label-agnostic one does trigger. This may be because of small per-label changes in confidence, yet significant changes when put together. Of course, as all the auditors are statistical, there are exceptions of all types. For example, instances in which a per-label auditor is superior.

\subsection{RQ 4: Effect of classifier ML performance on drift-detection ability}
For the MNIST data set we experiment not only with the well known neural network classifier model but also with a logistic regression model. These models differ in their ML performance. Specifically, the CNN model reaches 99\% accuracy while the logistic regression model only reaches 92\% accuracy. It is interesting to note that the less accurate the classifier the higher the drift percentage. That is, the weaker the classifier the weaker the drift detection. 
We saw this phenomenon in CIFAR10 as well, where the CNN model reaches 75\% accuracy. However, the logistic regression model had such poor accuracy that we did not include it in the results we report on here. A possible explanation is that a less-accurate model has a noisier confidence distribution and therefore provides less information to enable detection. At the extreme, a random classifier is likely to be completely robust to data drift.

\subsection{RQ 5: Auditor that is based on label distribution }
\begin{figure}[ht]
\caption{Outlier detection output on baseline (MNIST; left-out class 0)}
\label{fig:outlier}
\centering
\includegraphics[width=0.45\textwidth]{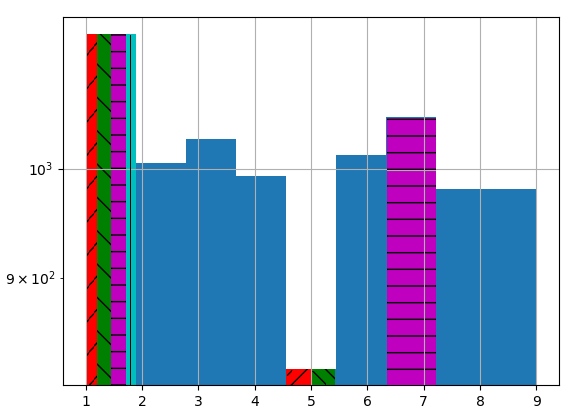}
\caption*{The figure shows the distribution of data records to classes for the baseline data. Classes (columns) with textures/colors other than plain blue were marked by one of the outlier methods. IQR (cyan/vertical lines) marked class 1; Hampel (magenta/horizontal lines) marked classes 1 and 7; z-score (green/backslash) marked classes 1 and 5; DB-scan (red/slash) marked classes 1 and 5.}
\end{figure}

We experimented with several common outlier detection algorithms: two quantile methods (IQR) \cite{outlier_iqr}, z-score \cite{mod_z_score}, Hampel \cite{hampel_outlier}, and DB-Scan \cite{dbscan}. 
The idea is to define an auditor that is based on the anomalous classification amount. The intuition is that often there exists a class that is the 'default' for previously unseen data. Figure \ref{fig:outlier} shows that all of these methods alerted on at least the highest and/or lowest values, even when no drift was introduced. This means that the false positive (FP) rate is very high which renders this approach useless. It is, of course, possible to tune any of these methods to capture the classification behavior as seen in training. However, this assumes knowledge of the class-wise distribution over input data records.

Our experimental results corroborate the fact that changes in label distribution do not necessarily indicate data drift. 

\begin{table*}[htbp]
  \centering
  \caption{Summary}
    \begin{tabular}{llllll}
    \toprule
    \textbf{Data Set} & \textbf{Model} & \textbf{Drift Type} & \textbf{Min \%} & \textbf{Max \%} & \textbf{Avg \%} \\
    \midrule
    \textbf{MNIST} & Logistic Regression & Type1 & 2  & 7  & 4.00 \\
    \textbf{MNIST} & Logistic Regression & Type3 & 5  & 12 & 8.00 \\
    \textbf{MNIST} & Logistic Regression & Type4 & 4  & 17 & 9.22 \\
    \textbf{MNIST} & Neural Network & Type1 & 1  & 2  & 1.11 \\
    \textbf{MNIST} & Neural Network & Type3 & 1  & 1  & 1.00 \\
    \textbf{MNIST} & Neural Network & Type4 & 1  & 21  & 4.33 \\
    \textbf{CIFAR10} & Neural Network & Type1 & 6  & 14* & 8.38 \\
    \textbf{CIFAR10} & Neural Network & Type4 & 7  & 9  & 7.55 \\
    \textbf{Kaggle loans} & Random Forest & Type1 &  1  & 3  & 1.57 \\
    \textbf{Kaggle loans} & Random Forest & Type4 &  1  & 1  & 1.00 \\
    \textbf{Kaggle loans} & Random Forest & Type2 & \textless1  &  & --- \\
    \bottomrule
    \end{tabular}%
  \label{tab:experimentSummary}%
\captionsetup{width=.9\linewidth}  
\caption*{Drift is simulated by inserting data records that differ from the training data. The drift type characterizes the difference. Type 1: held-out by class data records, Type 2:  held-out by category data records (top 10\% of loan\_amnt feature values in this case), Type 3: different data set records (CIFAR10 in this case), Type 4: randomly generated data records appropriate for the data domain. The percentage displayed in each cell is an upper bound (to the closest integer) on when drift was detected. (* model without 'car' never detects 'car' as drift)}
\end{table*}%

\begin{table}[htbp]
  \centering
  \caption{Logistic Regression model on MNIST (9 classes)}
  \begin{tabular}{p{1.25cm}p{1.5mm}p{1.5mm}p{1.5mm}p{1.5mm}p{1.5mm}p{1.5mm}p{1.5mm}p{1.5mm}p{1.5mm}p{1.5mm}l}
    \toprule
     & \multicolumn{10}{c}{Left-out class} & Avg \\
                 & 0  & 1  & 2  & 3  & 4  & 5  & 6  & 7  & 8  & 9  &  \\
    \midrule
    Type1 \% & 5  & 2  & 3  & 6  & 3  & 3  & 3  & 7  & 4  & 3  & 4.00 \\
    Type3 \% & 6  & 8  & 7  & 6  & 10 & 9  & 5  & 9  & 12 & 8  & 8.00 \\
    Type4 \% & 5  & 8  & 6  & 6  & 11 & 17 & 4  & 10 & 16 & 8  & 9.22 \\
    \bottomrule
    \end{tabular}%
  \label{tab:mnistlogreg}%
  \captionsetup{width=.95\columnwidth}  
  \caption*{  Summary for the different MNIST logistic regression models created when leaving out each of the label classes with different drift types. Type 1 drift is taken from the left-out class. Type 3 drift is taken from CIFAR10 and adjusted to the MNIST domain by reducing size and turning into gray scale. This results in pixels that display a centered image. Type 4 drift are random images create by randomly selecting each pixel gray scale.}
\end{table}%

\begin{table}[htbp]
  \centering
  \caption{Neural Network model on MNIST (9 classes)}
    \begin{tabular}{p{1.25cm}p{1.5mm}p{1.5mm}p{1.5mm}p{1.5mm}p{1.5mm}p{1.5mm}p{1.5mm}p{1.5mm}p{1.5mm}p{1.5mm}l}
    \toprule
     & \multicolumn{10}{c}{Left-out class} & Avg \\
                    & 0  & 1  & 2  & 3  & 4  & 5  & 6  & 7  & 8  & 9  &  \\
    \midrule
    Type1 \% & 1  & 1  & 1  & 1  & 2  & 1  & 1  & 1  & 1  & 1  & 1.1 \\
    Type3 \% & 1  & 1  & 1  & 1  & 1  & 1  & 1  & 1  & 1  & 1  & 1.0 \\
    Type4 \% & 21 & 2  & 8  & 1  & 3  & 1  & 1  & 1  & 1  & 1  & 4.0 \\
    \bottomrule
    \end{tabular}%
  \label{tab:mnistNN}%
  \captionsetup{width=.95\columnwidth}  
  \caption*{Summary for the different MNIST CNN models created when leaving out each of the label classes with different drift types. Drift types identical to Table \ref{tab:mnistlogreg}. }
\end{table}%

\begin{table}[htbp]
  \centering
  \caption{Neural Network model on CIFAR10 (9 classes)}
    \begin{tabular}{p{1.25cm}p{1.5mm}p{1.5mm}p{1.5mm}p{1.5mm}p{1.5mm}p{1.5mm}p{1.5mm}p{1.5mm}p{1.5mm}p{1.5mm}l}
        \toprule
     & \multicolumn{10}{c}{Left-out class} & Avg \\
     & \hd{airplane} & \hd{car} & \hd{bird} & \hd{cat} & \hd{deer} & \hd{dog} & \hd{frog} & \hd{horse} & \hd{ship} & \hd{truck} &  \\
    \midrule
    Type1 \% & 7  & 100 & 7  & 7 & 6  & 14 & 9  & 8 & 9  & 14 & 8.38 \\
    Type4 \% & 7  & 7   & 8  & 9 & 7  & 8  & 7  & 7  & 8  & 7 & 7.55 \\
    \bottomrule
    \end{tabular}%
    \captionsetup{width=.95\columnwidth}  
    \caption*{Summary for the different CIFAR10 CNN models created when leaving out each of the label classes with different drift types. Drift types similar to Table \ref{tab:mnistlogreg}.}
  \label{tab:cifar10NN}%
\end{table}%

\begin{table}[htbp]
  \centering
  \caption{Random Forest model on Kaggle loan data}
    \begin{tabular}{p{1.25cm}p{1.5mm}p{1.5mm}p{1.5mm}p{1.5mm}p{1.5mm}p{1.5mm}p{1.5mm}l}
    \toprule
     & \multicolumn{7}{c}{Left-out class} & Avg \\
             & 0  & 1  & 2  & 3  & 4  & 5  & 6   &  \\
    \midrule
    Type1 \% & 2  & 3  & 2  & 1  & 1  & 1  & 1   & 1.57 \\
    Type4 \% & 1  & 1  & 1  & 1  & 1  & 1  & 1   & 1.00 \\
    \bottomrule
    \end{tabular}%
  \label{tab:loanRF}%
\captionsetup{width=.95\columnwidth}  
\caption*{Summary of the different Kaggle loan data random forest models created for classifying into 'grade' (7 classes) when leaving out each of the label classes with different drift types. Type 1 drift uses the grade feature which is categorical but imbalanced, with grade\=6 having 100x less data points than other values. Type 2 drift inserts the top 10\% data records based on the loan\_amnt feature. 
}
\end{table}%

\subsection{RQ6: change-point analysis for statistical guarantees}
\label{cpm_evaluation}
To implement the sequential CPM method  we simulate sequences of observed data $x_1,x_2,\dots$, where each $x_i=p(c|d_i)$, the estimated class probability of the class $c$ that is predicted for a digit $d_i$, to demonstrate the statistical guarantees.  Given a model trained on a dataset with one digit class omitted, and with a fixed type of drift inserted, we randomly draw instances of $x$ with replacement from the test dataset, in sample sizes of size $n=20$. The first 50 samples of size 20 (that is, $\{x_1,\dots,x_{50\times 20}\}$) are instances of $p(c|d)$ where $d$ are digits the model was trained on, that is non-drift, so the drift time is $k=50\times20=1,000$.  Samples $51,\dots,70$ contain $1,2,\dots,20$ instances of drift (that is, observed $x_i=p(c|d_i)$ when $d_i$ is an image different than those trained on) randomly mixed with non-drift, and samples 71 onward have drift entirely.  The sampling is repeated 300 times to compare each method. 

Ideally, the distribution of prediction confidence will change when it encounters drift instances, and this mixture of drift and non-drift, observed indirectly through the confidences, will be detected by the CPM.  We insert the drift gradually, one additional instance per sample, because gradual change is harder to detect than sudden change in distribution.  If the sample $t$ during which drift is determined to have happened in the past, is $<51$, the determination is a false positive; the sooner after sample 50 that drift is detected, after observing some of the drift sequence, the more successful the CPM method is.  Along with several CPM sequential tests, such as Student-T and CvM, which sequentially control the false positive rate, we perform two non-sequential T-tests: 
\begin{itemize}
    \item ``Splits" T-test, which imitates the CPM technique of performing a two-sample T-test at every potential split after samples $k=1,\dots,t-1$ retroactively at each sample $t\geq 2$.
    \item ``Pairs" T-test, which conducts a two-sample T-test between sample 1 (which is non-drift) and each sample $t=2,\dots$.
\end{itemize}
These non-sequential T-tests are a naive approach to sequential detection because they use the traditional T-test critical values, which do not control the false positive rate properly, as shown in Table~\ref{tab:cpm_results} for an example where digit 0 was omitted, and out-of-domain (CIFAR) images were inserted as drift.  The CPM methods have low false positive rates (probability of deciding drift prior to sample 51), as desired.

\begin{table}[ht]
\centering
\caption{False positive results from CPM drift detection on data with digit 0 omitted and out-of-domain drift.}
\begin{tabular}{ll|rr}
  \hline
 Test & Type & Pr(Determined $k<51$) & Pr(Detection $t<51$) \\ 
  \hline
Student T& CPM & 0.247 & 0.157 \\ 
  Lepage &CPM &  0.097 & 0.017 \\ 
  CvM & CPM & 0.087 & 0.003 \\ 
\hline
  `Splits' T-test & non-seq & 0.000 & 0.000 \\ 
  `Pairs' T-test & non-seq & 0.580 & 0.580 \\ 
   \hline
\end{tabular}
\label{tab:cpm_results}
\end{table}

\begin{figure}[ht]
\caption{Distribution of time $t$ of detection of drift by CPM.}
\label{fig:distribution_cpm_detection_time}
\centering
\includegraphics[width=0.45\textwidth]{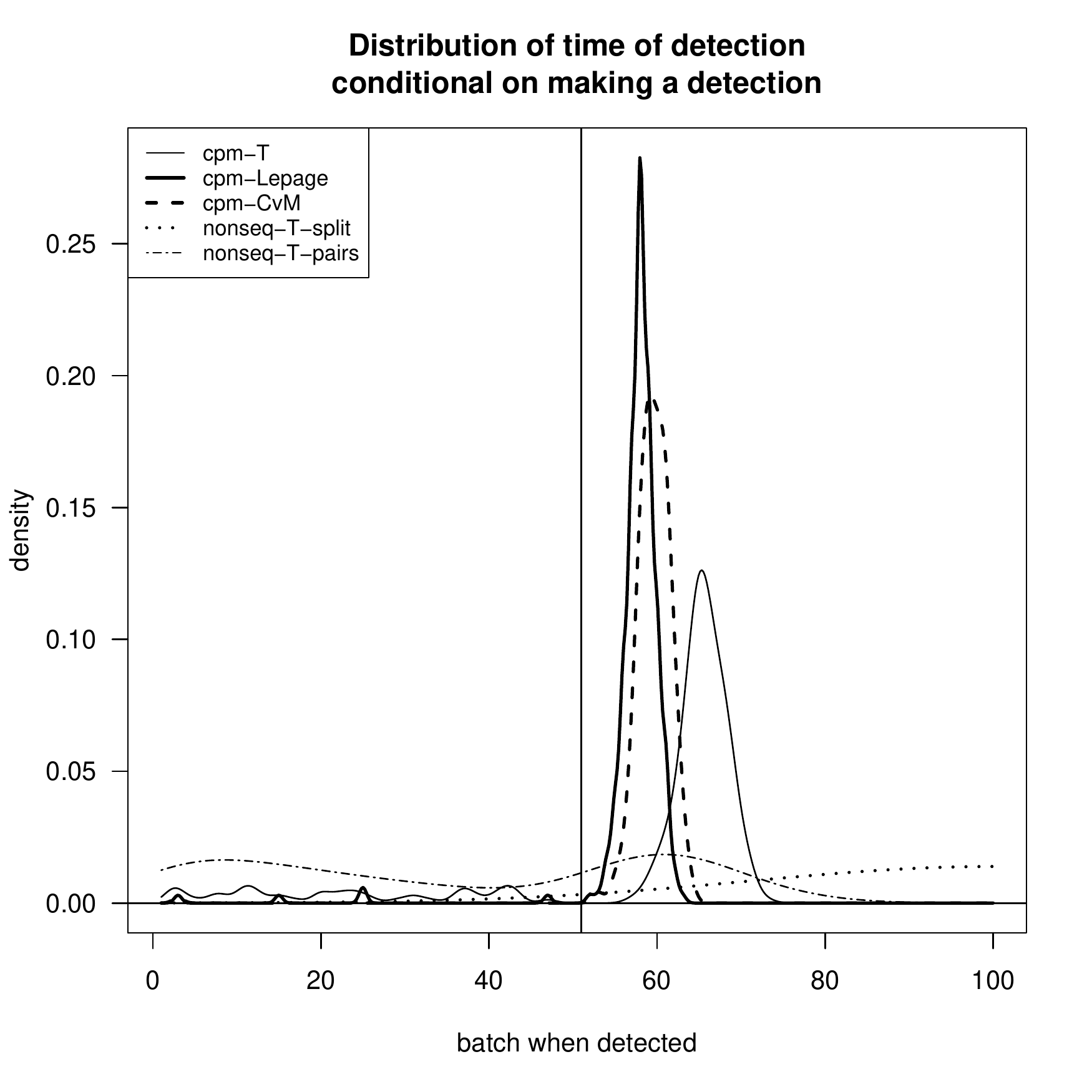}
\captionsetup{width=.95\columnwidth}  
\caption*{X axis specifies the sample $t$ during which drift was detected as happening in the past, with densities plotted over 300 data simulations.  The CPM methods all have distributions of $t$ concentrated shortly after the true change-point sample 50. The two non-sequential T-tests either frequently falsely detect drift before 50 (`pairs') or unacceptably late (`splits'). 
 }
\end{figure}

Figure~\ref{fig:distribution_cpm_detection_time} shows distribution of the drift detection time $t$ for the experiment from Table~\ref{tab:cpm_results}, over the 300 data simulations.  As Table~\ref{tab:cpm_results} shows, the CPM methods have low overall false positive probabilities (low density for $t\leq 50$) but also tend to detect drift shortly after it happens (high density close to, but after $t=50$).  This means they detect quickly but with few errors.  In this instance, the CPM T-test has more delayed detection than the more general LePage and CvM tests, likely if drift is not characterized by a change in the distribution mean.  The non-sequential `pairs' T-test has a high false positive rate (58\%), shown by a high density for $t\leq 50$.  The `splits' T-test, while having essentially no false positives, also has extremely delayed detection, with high density at the right edge of the plot.  Thus, these CPM experiments demonstrate that these methods can be applied to detect drift indirectly through the observed $p(c|d)$, with both quicker detection and lower false positive rates than non-sequential methods.

%% file: related.tex
\section{Related work} \label{sec:related}
Our work, and in particular the auditors that we define, goes beyond empirical distribution identification and comparison, to identify data drift. We want to alert on distribution changes that affect the ML performance. However, even the first problem is very difficult.

When the underlying class distribution is known analytically (e.g., parametric distribution or a mixture of distributions) then distribution comparison is straight forward. For example, using KL divergence \cite{klmeasure}. However, this is rarely the case in the real world. The data distribution is never perfectly known and therefore neither is the class distribution. One has to work with the empirical distribution which does not follow any specific parametric description of pre-defined distributions. Empirical density estimation has been well investigated and a variety of approaches exist for the classical settings in which the data has low dimensionality or is drawn from a known distribution or mixture of distributions. 
It has also been acknowledged that density estimation becomes very challenging for high-dimensional data \cite{Scholkopf:2001:ESH:1119748.1119749}. 
Modeling multivariate data with Bayesian networks is one example for a domain that has been struggling with this challenge \cite{SMITH2003387}. 
Recent advances propose high dimensionality density estimation, such as Maximum Mean Discrepancy \cite{Gretton:2012:KTT:2188385.2188410} or Wasserstein distance \cite{OLKIN1982257}. 
However, statistical calculations like a confidence interval, which depend on the distribution, become much more complicated for high dimension metrics than uni-variate metrics. Further, as the dimensionality increases so does the likelihood of having insufficient data for reliably performing such statistical calculations.

Our work investigates uni-variate statistical indicators---auditors---both for defining statistically significant data characteristics and to effectively identify when these break significantly (data drift). Working with uni-variate metrics has all the advantages of being able to apply classical statistics. Moreover, the amount of data needed is realistic. 

Other approaches try to improve the ML models to better handle changes in the data, identify data missing from the training examples or identify bias in the training examples. Such approaches often focus on handling missing data or direct learning to areas in the data where the model performs poorly.  Ensembles of weak learners such as AdaBoost  are an example.  Further research is required to apply these approaches to drift identification. Additional challenges exist for application over unlabeled production data. 

Various work captures challenges and best practices in developing ML solutions, e.g., \cite{Google_technical_debt,Google_testscore,Google_43rules}.
Our research addresses additional challenges that are not addressed by any of these existing approaches. Our research deals with capturing the expected envelope of operation of a ML model and monitoring for statistically significant changes against it. The approach is especially important when the production data is not labeled which is the case in many business applications. 

%% file: discussion.tex
\section{Discussion and conclusions}\label{sec:discussion}
We developed and implemented a framework for defining data drift detection auditors and for assessing their detection ability. We presented a novel auditor that is based on a classifier output---winning label and confidence in this label. Our experimental results, using three different data sets and multiple classifier models and drift types, demonstrate the effectiveness of this auditor for data drift detection. This requires neither production input data nor its ground truth. Our experiments further demonstrate that change-point analysis provides a promising means to statistically control the false-positive rate of drift detection auditors.

This is work in progress with many challenges and future directions. We are developing additional auditors that have complimentary data drift detection abilities.
We are working on enhancing our auditors framework and implementation.
 Many challenges exist in defining the end-to-end use-case of drift detection. For example, what should be done when an alert of data drift is triggered? One option is to label data from the production data to check the ML performance and potentially retrain or relearn. 